\documentclass[runningheads]{llncs}

\usepackage[review,year=2024,ID=5017]{eccv}
\usepackage{eccvabbrv}
\usepackage{graphicx}
\usepackage{amsmath,amssymb} % define this before the line numbering.
\usepackage{amsmath,amsfonts}
\usepackage{algorithmic}
\usepackage{algorithm}
\usepackage{array}
\usepackage{textcomp}
\usepackage{stfloats}
\usepackage{url}
\usepackage{bbm}
\usepackage{multirow}
\usepackage{verbatim}
\usepackage{threeparttable}
\usepackage{graphicx}
\usepackage{booktabs}
\usepackage{makecell}
\usepackage{diagbox}
\usepackage{arydshln}
\usepackage{color}
\usepackage{cite}
\usepackage{bm}
\usepackage{bbding}
\usepackage[width=122mm,left=12mm,paperwidth=146mm,height=193mm,top=12mm,paperheight=217mm]{geometry}
\usepackage{graphicx}
\usepackage{booktabs}

\usepackage[accsupp]{axessibility}  % Improves PDF readability for those with disabilities.

\usepackage[pagebackref,breaklinks,colorlinks,citecolor=eccvblue]{hyperref}
\usepackage{orcidlink}

\begin{document}

\title{Mixed Prototype Consistency Learning for Semi-supervised Medical Image Segmentation} 

\titlerunning{Abbreviated paper title}

\author{Lijian Li  \inst{1}\orcidlink{0009-0006-2009-3269} 
% \and
% Second Author\inst{2,3}\orcidlink{1111-2222-3333-4444} \and
% Third Author\inst{3}\orcidlink{2222--3333-4444-5555}
}


\institute{ Department of Computer and Information Science, University of Macau, Macau, China\\
\email{mc35305@umac.mo}
}

\maketitle

\begin{abstract}
Recently, prototype learning has emerged in semi-supervised medical image segmentation and achieved remarkable performance. However, the scarcity of labeled data limits the expressiveness of prototypes in previous methods, potentially hindering the complete representation of prototypes for class embedding. To address this problem, we propose the Mixed Prototype Consistency Learning (MPCL) framework, which includes a Mean Teacher and an auxiliary network. The Mean Teacher generates prototypes for labeled and unlabeled data, while the auxiliary network produces additional prototypes for mixed data processed by CutMix. Through prototype fusion, mixed prototypes provide extra semantic information to both labeled and unlabeled prototypes. High-quality global prototypes for each class are formed by fusing two enhanced prototypes, optimizing the distribution of hidden embeddings used in consistency learning. Extensive experiments on the left atrium and type B aortic dissection datasets demonstrate MPCL's superiority over previous state-of-the-art approaches, confirming the effectiveness of our framework. The code will be released soon.
\end{abstract}

\section{Introduction}
Recently, supervised medical image segmentation has achieved remarkable improvements by introducing deep learning methods. However, widely applying such techniques in real medical diagnosis is continually hindered by the scarcity of labeled data. Thus, researchers propose the concept of semi-supervised medical image segmentation to alleviate dependence of models on abundant manual annotations which require a significant amount of time and labor. SS-MIS methods are capable of achieving relatively great performance by extracting precious information from unlabeled data to assist model in training with a small amount of annotated data.

There are three common categories of semi-supervised learning techniques utilized in SS-MIS methods, including consistency-based, pseudo-label-based, and prototype-based learning approaches. As for pseudo-label-based methods \cite{DBLP:conf/cvpr/KwonK22, DBLP:conf/miccai/BaiOSSRTGKMR17, DBLP:conf/nips/ZhangWHWWOS21, DBLP:conf/cvpr/ChenYZ021}, they aim to generate pseudo-labels for unlabeled data to provide auxiliary supervisions similar to annotated data. Some studies introduce uncertainty estimation methods like Monte Carlo dropout \cite{yu2019uncertainty} and ensemble-based methods \cite{shi2021inconsistency} to mitigate interference information contained in unlabeled data so as to generate more reliable pseudo-labels. Besides, researchers also quantify the level of uncertainty and set a threshold to select high-confidence pseudo-labels by utilizing some measurements like information entropy. Xiang et al. \cite{xiang2022fussnet} considers both epistemic uncertainty (EU) and aleatoric uncertainty (AU) to generate a better uncertainty mask to promote model learning. Consistency-based methods \cite{DBLP:conf/nips/BachmanAP14, DBLP:conf/nips/SajjadiJT16, DBLP:conf/miccai/BortsovaDHKB19, DBLP:conf/icml/XuSYQLSLJ21, DBLP:journals/corr/abs-2001-04647, DBLP:conf/bmvc/FrenchLAMF20, DBLP:conf/aaai/LuoCSW21} comply with a learning pattern that guides model to generate consistent predictions for both raw data and enhanced or interfered data. A unified consistency loss is adopted to optimize model learning. Chen et al. \cite{DBLP:conf/cvpr/ChenYZ021} proposes a learning strategy named cross pseudo supervision (CPS) to enable pseudo labels generated by two perturbed segmentation networks to supervise mutually. Luo et al. \cite{DBLP:conf/aaai/LuoCSW21} introduces a dual-task consistency regularization (DTC) to enforce consistency between pixel-level and geometry-level segmentation maps for labeled and unlabeled data. However, slight or inappropriate perturbations may provide erroneous supervisory signals so as to produce suboptimal segmentation performance. In regard to prototype learning methods \cite{DBLP:conf/cvpr/WangWSFLJWZL22, DBLP:journals/pami/WuFHHMZ23, DBLP:conf/cvpr/Zhang0Z0WW21}, they are intended to impose a consistency constraint between prototypes generated by feature matching operations and corresponding model's predictions to improve segmentation performance. Xu et al. \cite{DBLP:journals/titb/XuWLYY000T22} devises a cyclic prototypical consistency learning framework (CPCL) that is composed of a labeled-to-unlabeled (L2U) prototypical forward process and an unlabeled-to-labeled (U2L) backward process. However, current prototype learning methods always generate prototypes for labeled and unlabeled data separately and fuse two kinds of prototypes to obtain more comprehensive prototypes. They are limited by the quantity and quality of prototypes so that the representation capability of global prototypes will be impaired. In response to the insufficient amount of annotated data, researchers usually use interpolation-based semi-supervised learning methods to solve this problem. For example, CutMix \cite{yun2019cutmix}, also known as Copy-Paste (CP), achieves data augmentation by copying image crops from labeled data as foreground onto another image. It has been verified that the predictions of mixed data can provide useful information to alleviate the distribution mismatch and class imbalance problems.

Briefly, previous prototype-based methods are hindered by the small quantity of prototypes and their low quality. To eliminate this problem, in this paper, we propose a Mixed Prototype Consistency Learning (MPCL) framework that introduces a Mean Teacher structure and an auxiliary network to enhance the prototype generation process. Firstly, the teacher and student networks are utilized to train unlabeled and labeled data respectively and generate corresponding prototypes by employing the masked average pooling approach \cite{wang2019panet}. The pseudo-labels for unlabeled data are optimized by an uncertainty estimation mechanism based on entropy. Then, we adopt CutMix \cite{yun2019cutmix} to generate mixed labeled data and devise an auxiliary network to generate auxiliary prototypes containing beneficial semantic information for mixed data. With auxiliary prototypes, labeled prototypes will be fused with them to enrich their common semantic information. Besides, we also fuse mixed prototypes with unlabeled prototypes to enhance semantic information interaction resulting in bridging the distribution gap between labeled and unlabeled data. Then, the enhanced labeled and unlabeled prototypes are fused to generate high-quality global prototypes, which have more powerful capabilities to represent the distribution of feature embeddings. Finally, we utilize cosine similarity to obtain the similarity maps between prototypes and feature embeddings for labeled or unlabeled data, which are regarded as novel segmentation masks and utilized to conduct consistency learning with labels and reliable pseudo-labels. In sum, the main contributions of this work are summarized as follows:
\begin{itemize}
    \item[1)] We introduce an auxiliary network to provide extra prototypes for mixed data formed by CutMix to assist model learning, which not only increases the number of prototypes model can consider but also improves the quality of global prototypes.

    \item[2)] A novel prototype-based framework named mixed prototype learning is proposed, which utilizes extra prototypes to enrich semantic information of labeled and unlabeled prototypes so as to generate high-quality global prototypes to optimize the distribution of hidden embeddings employed in consistency learning.

    \item[3)] The results of extensive experiments conducted on Left Atrium and Aortic Dissection datasets verify that the proposed MPL framework outperforms previous SOTA SSL methods
\end{itemize}

\section{Related works}
\textbf{Semi-supervised Medical Image Segmentation}
Recently, a large quantity of semi-supervised learning methods have been introduced into the field of medical image segmentation and achieved quite considerable breakthroughs. The existing semi-supervised medical image segmentation methods generally utilize an encoder-decoder segmentation network like V-Net as their backbone and mainly focus on devising an excellent learning strategy. Consistency learning is one of the most commonly used strategies. Most consistency learning-based methods adopt the structure of Mean Teacher and design various regularization methods to maintain the consistency of pseudo-labels generated by teacher and student networks. For example, Yu et al. \cite{yu2019uncertainty} applies an uncertainty-guided learning strategy based on the framework of Mean Teacher to enable student network to learn more reliable targets, which are optimized by some uncertainty estimation methods like Monte Carlo Dropout. Wu et al. \cite{DBLP:conf/miccai/WuXGCZ21} devise a mutual consistency network (MC-Net) to mitigate uncertain information that model brings through imposing a consistency constraint on two predictions generated by two identical decoders. Xia et al. \cite{DBLP:journals/mia/XiaYYLCYZXYR20} attempts to enforce the consistency of predictions for multi-view unlabeled data. Ouali et al. \cite{ouali2020semi} proposes a cross consistency training strategy that enforces consistency between the predictions of main decoder and multiple auxiliary decoders whose inputs are processed by some perturbations. Besides, Luo et al. \cite{luo2021efficient} devises a novel uncertainty rectifying strategy to ensure the consistency of pyramid predictions at different scales. Zhu et al. \cite{zhu2023hybrid} utilizes a hybrid uncertainty to eliminate unreliable knowledge in the hybrid prediction and leverage beneficial information to assist model learning. Recently, prototype alignment based on feature matching operations has been applied in semi-supervised medical image segmentation, which aims to obtain data structure information from both labeled and unlabeled data and improve the distribution of feature embeddings across distinct classes. For instance, Zhang et al. \cite{zhang2023self} proposes a self-aware and cross-sample prototypical learning network (SCP-Net) to leverage semantic interaction information from two different prototypes and a self-aware consistency loss to improve the reliability of model. Wu et al. \cite{wu2022exploring} enhances the cohesion of each category distribution by aligning the features of specific categories with their high-quality prototypes, thereby achieving better separation of categories. Zhang et al. \cite{DBLP:conf/cvpr/Zhang0Z0WW21} proposes a progressively self-correcting strategy for pseudo-labels that calculates the feature distances from prototypes and leverages them as class-wise likelihoods to correct pseudo-labels. Zhang et al. \cite{zhang2020sg} introduces the masked average pooling strategy to obtain prototypes of images and utilizes cosine similarity to conduct pixel-level feature-to-prototype matching. Adversarial learning is also an advanced technique that has been widely applied in medical image segmentation. For example, Peiris et al. \cite{peiris2021duo} combines adversarial learning and multi-view learning to obtain confident prediction masks for unlabeled data. Wang et al. \cite{wang2023cat} devises a Constrained Adversarial Training (CAT) method which introduces various constraints like connectivity and convexity to generate anatomically plausible segmentations.

\textbf{Interpolation-based Semi-supervised Learning}
To address the problem of the scarcity of labeled data, interpolation-based regularization is applied to semi-supervised semantic segmentation and achieves tremendous success. Mixup \cite{zhang2017mixup} and CutMix \cite{yun2019cutmix} are two fundamental tasks that assist model training by mixing entire images and mixing image crops, respectively. There are numerous variants designed for specific problems based on them. For example, Tu et al. \cite{tu2022guidedmix} leverages mixup to transfer knowledge from labeled data into unlabeled data so as to generate more reliable pseudo labels. Dvornik et al. \cite{dvornik2018modeling} employs visual context information to find appropriate locations to place new objects on images so as to perform data augmentation. Bai et al. \cite{bai2023bidirectional} proposes a bidirectional copy-paste (BCP) strategy to mix labeled and unlabeled crossly to enable student network to learn more conjoint semantic information, which is regarded as mixed supervisory signals to assist model learning. Berthelot et al. \cite{berthelot2019mixmatch} gradually expands the annotated set by regarding unlabeled data with high-quality pseudo-labels as annotated data.

\section{Methodology}
\subsection{Problem Definition}
\begin{figure}
    \centering
    \includegraphics[scale=0.71]{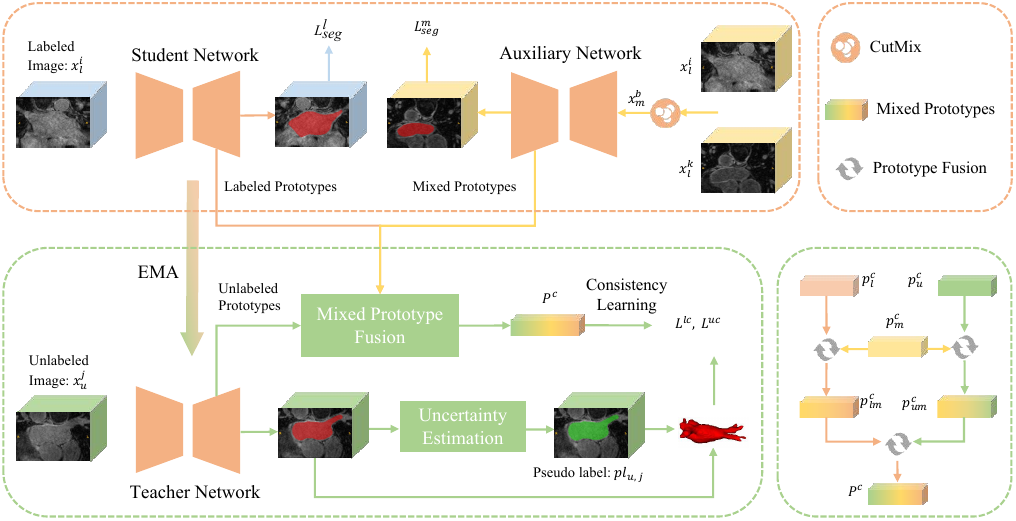}
    \caption{The figure illustrates the structure of the proposed Mixed Prototype Consistency Learning framework, which is composed of a Mean Teacher structure and an auxiliary network. The student and teacher networks are utilized to generate labeled and unlabeled prototypes $p^c_l, p^c_u$, respectively. And the auxiliary network produces mixed prototypes $p^c_m$ for mixed data processed by CutMix. The mixed prototype fusion module includes three fusion processes. The labeled and unlabeled prototypes will be fused with mixed prototypes to enhance their semantic information. Finally, a high-quality global prototype representation $p^c$ is formed by fusing labeled and unlabeled prototypes, which optimizes the distribution of hidden embeddings in consistency learning.}
    \label{structure}
\end{figure}

In the task of semi-supervised medical image segmentation, we have a whole dataset $\mathbb{D}$ including an annotated subset and an unannotated subset, which are defined as $\mathbb{D}_l, \mathbb{D}_u$ respectively. The annotated subset consists of $N$ images and corresponding labels denoted as $\mathbb{D}_l = \{(x^a, y^a)\}_{a=1}^N$, and the unlabeled subset only contains $M$ images denoted as $\mathbb{D}_u = \{(x^a)\}_{a=N+1}^{N+M}$. Besides, the dimensions of each image are $H \times W \times D$ which represent height, width and depth. Therefore, we denote a single image as $x^a \in \mathbb{R}^{H \times W \times D}$. For each label, it is defined as $y^a \in \{0,1,...,C-1\}^{H \times W \times D}$. The batch size for both labeled and unlabeled is 2, which is denoted as $\mathcal{B}$.The proposed mixed prototype learning framework is composed of a Mean Teacher structure and an auxiliary network, whose overview structure is illustrated in Figure \ref{structure}. The student network and auxiliary network are optimized by optimizers like stochastic gradient descent (SGD). Then, the parameters of teacher network are updated by exponential moving average (EMA) of student network.
\subsection{Supervised Loss}
To simply mitigate the uncertainty of predictions for labeled data that model brings, we employ four different loss functions, such as focal loss, Dice loss, cross entropy loss and IoU loss, to process different outputs from an identical group of classifiers. Besides, different from \cite{xiang2022fussnet}, we also utilize cross entropy loss function to process the average of four distinct predictions to further eliminate disparity between the four classifiers' results. Therefore, the supervised loss for labeled data is the sum of average of four different losses and the fused cross entropy loss, which is defined as follows:
\begin{equation}
    L_{seg}^l = (L_{ce}^l + L_{dice}^l + L_{focal}^l + L_{IoU}^l) / 4 + L^l_f 
\end{equation}
where $L_{ce}, L_{dice}, L_{focal}, L_{IoU}$ represent cross entropy loss, dice loss, focal loss and IoU loss for labeled data respectively and $L_{f}$ denotes the fused cross entropy loss. For the auxiliary network, identical supervised loss functions are constructed to train mixed data processed by CutMix which is to copy crop from one image onto another image. Therefore, the mixed data is defined as $\mathbb{D}_m = \{(x_m^b, y_m^b)\}_{b=1}^{N/2}$ where the number of mixed images is $N/2$. 
\begin{equation}
    L_{seg} = L_{seg}^l + L_{seg}^m
\end{equation}
where $L_{seg}$ denotes the final supervised loss and $L_{seg}^m$ represents the supervised loss for mixed data. 

The pseudo-label for the $p$-th voxel in the $a$-th unlabeled image is denoted as $pl_{u,a}^p \in \mathcal{R}^C$. Similarly, the average of multiple different predictions for unlabeled images will be utilized to generate pseudo labels $pl_{u}$. To progressively improve the reliability of pseudo-labels, we adopt uncertainty measurement based on entropy to estimate the quality of pseudo-label for each voxel. According to the properties of entropy, the lower entropy of a voxel, the lower its uncertainty which also means that the voxel is more reliable. The uncertainty for $p$-th voxel is defined as follows:
\begin{equation}
    U(pl_{u,a}^p) = -\sum_{c=0}^{C-1}pl_{u,a}^p(c)\log pl_{u,a}^p(c)
\end{equation}

With the uncertainty map, reliable map for each voxel can be naturally inferred which is regarded as weight assigned to each voxel. Therefore, the reliable pseudo labels for unlabeled data can be obtained by multiplying the reliable map with raw pseudo labels. The process can be defined as follows:
\begin{equation}
    \hat{pl}_{u,a}^p = \frac{1}{H\times W\times D} (1 - \frac{U(pl_{u,a}^p)}{\sum_{p=1}^{H\times W \times D}U(pl_{u,a}^p)})\otimes pl_{u,a}^p
\end{equation}
where $\hat{pl}_{u,a}^p$ represents reliable pseudo-label for $p$-th voxel, and $\otimes$ denotes element-wise multiplication operation.
\subsection{Mixed Prototype Learning}
\textbf{Prototype Generation.} Different from utilizing aforementioned predictions, prototypes for labeled and unlabeled data are generated by feature embeddings from the $k$-rd layer decoder of student and teacher networks separately. Through trilinear interpolation, prototypes are upsampled to the same size as segmentation labels. Here, we set the feature maps produced by the $k$-rd layer of student network and teacher network to be $f_{k, a}^{l, p}$ and $f_{k, a}^{u, p}$. Referencing \cite{wang2019panet}, we directly utilize labels to mask feature maps and then adopt masked average pooling to obtain the prototype for each class. The labeled prototype for class $c$ is defined as follows:
\begin{equation}
    p^c_l = \frac{1}{\mathcal{B}}\sum_{a=1}^\mathcal{B} \frac{\sum_{p=1}^{H\times W \times D} f_{k, a}^{l, p} \mathbbm{1}[y^a_p=c]}{\sum_{p=1}^{H\times W \times D}\mathbbm{1}[y^a_p=c]}
\end{equation}

Different from the generation of labeled prototypes, we utilize uncertainty measurement of each voxel to optimize feature maps $f_{k, a}^{u, p}$. Before conducting the operation of masked attention pooling, the feature maps are weighted by corresponding uncertainty maps. The unlabeled prototype for class $c$ is formulated as follows:
\begin{equation}
    p^c_u = \frac{1}{\mathcal{B}}\sum_{a=1}^\mathcal{B}  \frac{1}{H\times W\times D}\frac{\sum_{p=1}^{H\times W \times D}  (1 - \frac{U(f_{k, a}^{u, p})}{\sum_pU(f_{k, a}^{u, p})}) f_{k, a}^{u, p}  \mathbbm{1}[y^a_p=c]}{\sum_{p=1}^{H\times W \times D}\mathbbm{1}[y^a_p=c]}
\end{equation}

Due to the limited annotated data, the annotated prototypes cannot fully express the feature embeddings of specific classes. Considering the problem, we utilize an auxiliary network to process mixed data and produce mixed prototypes according to the prototype generation method for labeled data. The feature maps from auxiliary network are represented as $f_{k, b}^{m, p}$. And the batch size for mixed data is $\mathcal{B}^m = \frac{\mathcal{B}}{2}$. The definition of mixed prototype $p^c_m$ is as follows:
\begin{equation}
    p^c_m = \frac{1}{\mathcal{B}^m}\sum_{b=1}^{\mathcal{B}^m} \frac{\sum_{p=1}^{H\times W \times D} f_{k, b}^{m, p} \mathbbm{1}[y^b_{m, p}=c]}{\sum_{p=1}^{H\times W \times D}\mathbbm{1}[y^b_{m, p}=c]}
\end{equation}
We utilize the mixed prototypes as additional knowledge to enhance the semantic information of both labeled and unlabeled prototypes.

\textbf{Mixed Prototype Fusion.}
For the fusion of different categories of prototypes, we first fuse mixed prototypes with labeled prototypes to provide labeled prototypes with more semantic information to enhance their capabilities to express class embeddings. The fusion process is defined as follows:
\begin{equation}
    p^c_{lm} = \lambda_1 p^c_l + \lambda_2 p^c_m
\end{equation}
where $p^c_{lm}$ represents the prototypes generated after the fusion of mixed and labeled prototypes, and $\lambda_1$ and $\lambda_2$ are the coefficients to adjust the fusion ratio of mixed and labeled prototypes. 

In addition, we will also integrate mixed prototypes with unlabeled prototypes to enable model to learn more common semantics from labeled and unlabeled data, which bridges the huge distribution gap between both data and promotes distribution alignment. 
% Similarly, we regard unlabeled prototypes as the primary semantic source and gradually extract knowledge from mixed prototypes to enhance unlabeled prototypes. 
% \begin{equation}
%     p^c_{um} = ((2 - \lambda_{con})p^c_u + \lambda_{con}p^c_m) / 2
% \end{equation}
The unlabeled prototypes enhanced by mixed prototypes are denoted as $p^c_{um}$. 
\begin{equation}
    p^c_{um} = \lambda_3 p^c_u + \lambda_4 p^c_m
\end{equation}
where $\lambda_3$ and $\lambda_4$ denote the coefficients for the fusion of unlabeled and mixed prototypes. Then, we fuse the optimized labeled and unlabeled prototypes to generate high-quality global prototypes $p^c$, which further bridge the distribution gap between labeled and unlabeled data and have better expression ability for feature embeddings. We adopt Temporal Ensembling approach that is capable of gradually improving the quality of predictions and prototypes during the training process to progressively update the fusion percentage of distinct prototypes. Time-dependent Gaussian warming up function \cite{tarvainen2017mean} is a prevalent technique that includes a vital parameter $\lambda_{con}$ ranging from 0 to 1. As the training progresses, the coefficients for labeled and unlabeled prototypes enable model to continually focus on the mixed prototypes while maintaining labeled prototypes as the capital information sources.
\begin{equation}
    p^c = ((2 - \lambda_{con})p^c_{lm} + \lambda_{con}p^c_{um}) / 2
\end{equation}

\subsection{Loss functions}
With the global prototypes, the next step is to build relationships between prototypes and features from teacher and student networks. We utilize cosine distance to measure the similarity \cite{zhang2020sg}. Then, the likelihood of voxels in each class is approximated using feature-to-prototype similarity maps.
% \begin{equation}
%     s_a^p = Sim(f_{k, a}^p, p^c)=\frac{f_{k, a}^p \cdot p^c}{max(\Vert f_{k, a}^p\Vert_2 \cdot \Vert p^c \Vert_2, \theta)}
% \end{equation}
% where $\theta$ is set to $1e^{-8}$, $Sim$ denotes cosine similarity and $s_a$ are prototype-based predictions for all voxels in $a$-th image. 
To conduct consistency learning, we utilize cross entropy loss function to narrow the distance between prototype-based predictions for labeled data and labels. For prototype-based predictions for unlabeled data, they are regarded as beneficial information and are expected to close to pseudo labels $\hat{pl}_{u,a}^p$ to improve their qualities. The consistency losses based on prototypes for labeled and unlabeled data are defined as follows:
\begin{equation}
    L_{lc} = L_{ce}(s_{l, a}, y_a), \qquad  L_{uc} = \sum_{p=1}^{H\times W \times D} L_{ce}(s_{u, a}^p, \hat{pl}_{u, a}^p)
\end{equation}
where $s_a^l$ and $s_a^u$ denote the similarity maps between labeled prototypes and features for labeled and unlabeled data respectively. Finally, the total loss of the proposed mixed prototype learning is combined with all supervised losses and two consistency losses, which is shown in below equation:
\begin{equation}
    L = L_{seg} + L_{lc} + \lambda_{con}L_{uc}
\end{equation}
\section{Experiments}
\subsection{Datasets and Metrics}
Two public datasets are utilized to evaluate the performance of our model, including the left atrium (LA) dataset \cite{DBLP:journals/mia/XiongXHHBZVRMYH21} and a multi-center dataset for type B aortic dissection (TBAD). For both datasets, we normalize the voxel intensities to zero mean and unit variance.

\textbf{LA dataset} \cite{DBLP:journals/mia/XiongXHHBZVRMYH21} is comprised of 100 3D gadolinium-enhanced magnetic resonance (MR) imaging volumes, each of which has identical spatial resolution of $0.625 \times 0.625 \times 0.625$\(mm^3\) and distinct dimensions. The dataset is methodically partitioned to allocate 80 samples for the training set and 20 samples for the validation set. For training, samples in training set are randomly cropped into $112 \times 112 \times 80$ patches. For inference, a sliding window of the same dimensions is employed, along with a stride of $18 \times 18 \times 4$, to generate the final segmentation outcomes. 

\textbf{TBAD dataset} \cite{yao2021imagetbad} consists of 124 computed tomography angiography (CTA) scans having three annotations (whole aorta, true lumen (TL), and false lumen (FL)). In the dataset, 100 and 24 scans are utilized for training and testing, respectively. Besides, for the preprocessing of data, we utilize the same techniques as in \cite{lu2023upcol}.

\textbf{Metrics.} We evaluate our model with four common metrics: Dice coefficient (Dice), Jaccard Index (Jac), 95\% Hausdorff Distance (95HD), and Average Symmetric Surface Distance (ASD). Two former and latter metrics are regionally sensitive and edge-sensitive, respectively. 

\subsection{Implementation details}
We employ V-Net \cite{milletari2016v} as backbone. And the student network and auxiliary network are trained for 10k iterations and optimized by an Adam optimizer with a learning rate of 0.01, while the parameters of teacher network are updated using exponential moving average (EMA). The batch size is set to 4, containing 2 annotated and unannotated samples. During the training stage, we augment the TBAD dataset with the same approaches like rotation following \cite{lu2023upcol} and use 3-fold cross-validation. And $k$ is set to 1, which is the optimal number of decoder layer to generate prototypes in student, teacher and auxiliary networks. Besides, the fusion coefficients for labeled prototypes $\lambda_1$ and $\lambda_2$ are equal. The same setting also appears on the fusion coefficients of unlabeled prototypes $\lambda_3$ and $\lambda_4$. The default loss function utilized in consistency learning is cross entropy. CutMix is set to be the default data augmentation technique. The whole MPCL framework is implemented by PyTorch and trained with an RTX 4090 GPU.

\begin{table}[htbp]\small
\centering
\caption{Experimental Results Comparison on the LA Dataset}
  \setlength{\tabcolsep}{0.8mm}{\begin{tabular}{c|cc|cccc}
    \hline
    \multirow{2}*{Method} &\multicolumn{2}{c|}{Scans Used} &\multicolumn{4}{c}{Metrics}\\
    \cline{2-7}
    {}&{Labeled} & {Unlabled} & Dice$\uparrow$ & Jaccard$\uparrow$ & 95HD$\downarrow$ & ASD$\downarrow$ \\
    \hline
    V-Net & 4(5\%) & 0 & 52.55 & 39.60 & 47.05 & 9.87 \\
    V-Net & 8(10\%) & 0 & 82.74 & 71.72 & 13.35 & 3.26 \\
    V-Net & 80(All) & 0 & 91.47 & 84.36 & 5.48 & 1.51 \\
    \hline
    UA-MT & \multirow{8}*{16(20\%)} &\multirow{8}*{64(80\%)} & 88.88 & 80.21 & 7.32 & 2.26 \\
    SASSNet  &{}&{}& 89.54 & 81.24 & 8.24 & 1.99 \\
    DTC  &{}&{}& 89.42 & 80.98 & 7.32 & 2.10 \\
    URPC &{}&{}& 88.43 & 81.15 & 8.21 & 2.35 \\
    MC-Net &{}&{}& 90.12 & 82.12 & 11.28 & 2.30 \\
    SS-Net &{}&{}& 89.25 & 81.62 & 6.45 & 1.80 \\
    BCP &{}&{} &90.34 & 82.50 & 6.75 & 1.77 \\
    Co-BioNet &{}&{}& 91.26 & 83.99 & 5.17 & 1.64 \\
    MPCL&{}&{} & \textbf{91.98} & \textbf{85.02} & \textbf{4.77} & \textbf{1.58}\\
    \hline
  \end{tabular}}
  \label{LA}
\end{table}

\begin{table}[htbp]\tiny
    \renewcommand{\arraystretch}{1.6}
	\label{table5}
	\centering
        \caption{Comparison with state-of-the-art models on Aortic Dissection dataset}
			\setlength{\tabcolsep}{0.8mm}{
                \begin{tabular}{c |c c |c c c |c c c| c c c |c c c  }
				\hline
                    \multicolumn{1}{c|}{\multirow{2}*{Dataset}}& \multicolumn{2}{c|}{\multirow{2}*{Scans Used}} &\multicolumn{12}{c}{Metrics}\\\cline{4-15}
				{}&{} &{} &\multicolumn{3}{c|}{Dice$\uparrow$}&\multicolumn{3}{c|}{Jaccard$\uparrow$}&\multicolumn{3}{c|}{95HD$\downarrow$}&\multicolumn{3}{c}{ASD$\downarrow$}  \\\cline{1-15}
                \multicolumn{1}{c|}{\multirow{1}*{Model}}&{Labeled} & {Unlabled} & TL & FL & Mean & TL & FL & Mean & TL & FL & Mean& TL & FL & Mean \\ \cline{1-15}
                \multirow{2}*{V-Net}&(20\%)&(80\%)&55.51& 48.98 &52.25&39.81 &34.79& 37.30 &7.24& 10.17& 8.71& 1.27 &3.19 &2.23 \\
                &(100\%)&(0\%)&  75.98& 64.02& 70.00& 61.89& 50.05& 55.97& 3.16& 7.56& 5.36 &0.48& 2.44 &1.46 \\\hline
				MT&\multirow{5}*{(20\%)} & \multirow{5}*{(80\%)} &  57.62& 49.95& 53.78& 41.57& 35.52&38.54& 6.00& 8.98& 7.49& 0.97& 2.77& 1.87 \\
				UA-MT&&&70.91& 60.66 &65.78& 56.15& 46.24& 51.20& 4.44& 7.94& 6.19& 0.83& 2.37& 1.60  \\
				FUSSNet&&&  79.73& 65.32& 72.53& 67.31& 51.74& 59.52& 3.46& 7.87& 5.67& 0.61& 2.93& 1.77  \\
				URPC&&&  81.84& 69.15& 75.50& 70.35& 57.00& 63.68& 4.41& 9.13& 6.77& 0.93& 1.11 &1.02  \\
				UPCoL&&&  82.65& 69.74& 76.19& 71.49& 57.42& 64.45& 2.82& 6.81& 4.82& 0.43& 2.22& 1.33 \\ \hline
			MPCL&(20\%)&(80\%)&\textcolor{red}{\textbf{83.83}}& \textcolor{red}{\textbf{76.05}} & \textcolor{red}{\textbf{79.94}} & \textcolor{red}{\textbf{73.00}}&\textcolor{red}{\textbf{63.63}}&\textcolor{red}{\textbf{71.94}}&\textcolor{red}{\textbf{2.24}}&\textcolor{red}{\textbf{5.02}}&\textcolor{red}{\textbf{3.63}} & \textcolor{red}{\textbf{0.33}} & \textcolor{red}{\textbf{1.15}} & \textcolor{red}{\textbf{0.74}}\\
                \hline
		\end{tabular}}
 \label{1}
\end{table}

\begin{figure}
    \centering
    \includegraphics[scale=0.23]{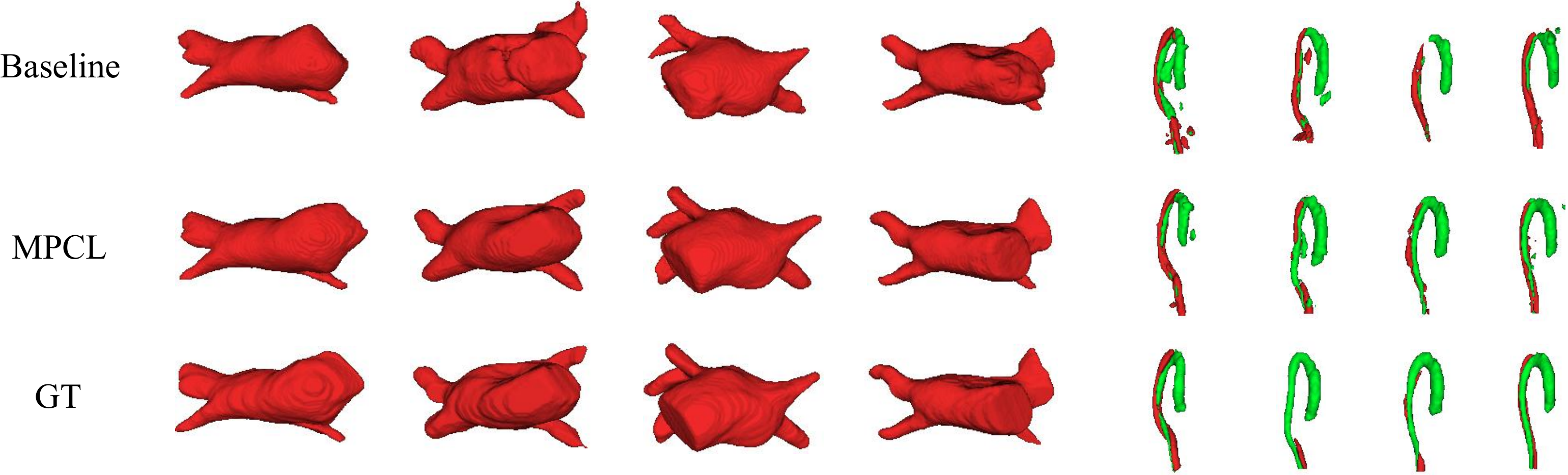}
    \caption{The visualizations of experimental results on Left Atrium and Aortic Dissection datasets. MPCL denotes the proposed mixed prototypes consistency learning. GT denotes the ground truth labels. Baseline represents the method of URPC. ITK-SNAP \cite{py06nimg} is the tool to visualize 3D medical image.}
    \label{vis}
\end{figure}

\subsection{Discussions}
\textbf{Results on Left Atrium dataset.}
The table \ref{LA} presents a comparison of various state-of-the-art models on the Left Atrium dataset, focusing on their performance with a labeled ratio of 20\%. The models compared are UA-MT \cite{DBLP:conf/miccai/YuWLFH19}, SASSNet \cite{DBLP:conf/miccai/LiZH20}, DTC \cite{DBLP:conf/aaai/LuoCSW21}, URPC \cite{luo2022semi}, MC-Net \cite{DBLP:conf/miccai/WuXGCZ21}, SS-Net \cite{wu2022exploring}, Co-BioNet \cite{peiris2023uncertainty}, BCP \cite{bai2023bidirectional}. Obviously, across all metrics (Dice, Jaccard, 95HD, ASD), the proposed model consistently outperforms previous state-of-the-art models on the Left Atrium dataset. For a 20\% labeled ratio, the proposed MPCL attains the highest Dice score (91.98) and Jaccard index (85.02), along with the lowest 95HD (4.77) and ASD (1.58) among all models compared. In order to better observe the effectiveness improvement of the proposed MPCL framework, we visualize the predictions of MPCL on LA dataset and compare them with the predictions of URPC model and ground truth labels in Fig. \ref{vis}. Obviously, the visualization results of MPCL's predictions are far superior to those of baseline's predictions, with a very high similarity to corresponding labels. To be specific, for the first and third examples in Fig. \ref{vis}, the predictions of baseline lack a section of content compared with the ground truth labels. However, the predictions of MPCL effectively compensate for the missing parts, with only slight differences in overall shape from the label, which reflects the improvements of MPCL on Dice and Jaccard score metrics.

\textbf{Results on Aortic Dissection dataset.}
In the Aortic Dissection dataset comparison illustrated in table \ref{1}, the proposed MPCL framework also excels, showing remarkable performance improvements at lower labeled data scenarios (like 20\% labeled data) compared with previous methods including MT \cite{tarvainen2017mean}, UA-MT \cite{DBLP:conf/miccai/YuWLFH19}, FUSSNet \cite{xiang2022fussnet}, URPC \cite{luo2022semi} and UPCoL \cite{lu2023upcol}. MPCL achieves the highest Dice (up to 79.94\%), Jaccard (up to 71.94\%), and significantly lower 95HD (down to 3.63) and ASD (down to 0.74) metrics, showing robust capability of capturing sophisticated geometric information like the vessel walls between True Lumen (TL) and False Lumen (FL). Similarly, we also visualize the predictions of MPCL and baseline on Aortic Dissection dataset and compare them with ground truth labels in Fig. \ref{vis}. According to the visualization results, baseline is inclined to predict an incomplete structure or generate redundant parts, which can be observed in the five and seventh examples in Fig. \ref{vis}. Compared with baseline, the predicted shape of MPCL appears more aggregated and more consistent with the labels. Besides, the accuracy of category prediction of MPCL is also higher based on the color distribution of the predicted images.

\subsection{Ablation studies}
We conduct extensive ablation studies to explore the effectiveness of each component of the proposed framework and various optimal parameter settings. There are five distinct ablation studies targeting data augmentation techniques, prototype fusion steps, prototype fusion coefficients, parameter k, and loss functions used in consistency learning.

\begin{table}[htbp]\small
\centering
\caption{Ablation studies of data augmentation techniques used for generation of mixed data on the LA Dataset}
  \setlength{\tabcolsep}{0.8mm}{\begin{tabular}{c|cc|cccc}
    \hline
    \multirow{2}*{Method} &\multicolumn{2}{c|}{Scans Used} &\multicolumn{4}{c}{Metrics}\\
    \cline{2-7}
    {}&{Labeled} & {Unlabled} & Dice$\uparrow$ & Jaccard$\uparrow$ & 95HD$\downarrow$ & ASD$\downarrow$ \\
    \hline
    CutMix & 8(20\%) & 0 & \textbf{91.98} & \textbf{85.02} & \textbf{4.77} & 1.58 \\
    Mixup & 8(20\%) & 0 & 91.01 & 83.62 & 5.27 & 1.78 \\
    Cutout & 8(20\%) & 0 & 91.58 & 84.52 & 5.00 & \textbf{1.52} \\
    fMix & 8(20\%) & 0 & 91.44 & 84.29 & 5.06 & 1.72 \\
    \hline
  \end{tabular}}
  \label{data}
\end{table}

As aforementioned, we utilize CutMix as the default data augmentation method for the generation of mixed prototypes. Here, we attempt to explore the impact of different data augmentation methods on model performance. The table \ref{data} presents the results of an ablation study that evaluates the impact of different data augmentation techniques on LA dataset. The study focuses on four augmentation methods: CutMix, Mixup, Cutout \cite{devries2017improved}, and fMix \cite{harris2020fmix}. CutMix appears to be the most effective data augmentation technique for this dataset, achieving the highest Dice and Jaccard scores (91.98\% and 85.02\%, respectively) and the lowest Hausdorff Distance at 4.77. Cutout results in the lowest ASD score (1.52), indicating that mixed data processed by Cutout is capable of providing edge-related semantic information to global prototypes that enable model to achieve more precise surface delineation. Mixup and fMix show competitive performance but lag slightly behind CutMix and Cutout in these evaluations.

\begin{table}[htbp]\small
\centering
\caption{Ablation studies of prototype fusion steps on the LA Dataset. N-N denotes that the fusion processes related to mixed prototypes are removed. M-L represents that the fusion of mixed and labeled prototypes is removed. M-UL denotes that the fusion of mixed and unlabeled prototypes is removed. L-UL represents that the fusion of labeled and unlabeled prototypes is removed.}
  \setlength{\tabcolsep}{0.8mm}{\begin{tabular}{c|cc|cccc}
    \hline
    \multirow{2}*{Method} &\multicolumn{2}{c|}{Scans Used} &\multicolumn{4}{c}{Metrics}\\
    \cline{2-7}
    {}&{Labeled} & {Unlabled} & Dice$\uparrow$ & Jaccard$\uparrow$ & 95HD$\downarrow$ & ASD$\downarrow$ \\
    \hline
    N-N & 8(20\%) & 0 & 90.78 & 82.96 & 8.72 & 3.05 \\
    M-L & 8(20\%) & 0 & 91.38 & 84.22 & 7.82 & 2.46 \\
    M-UL & 8(20\%) & 0 & 91.32 & 84.15 & 7.56 & 2.33 \\
    L-UL & 8(20\%) & 0 & 91.07 & 83.86 & 8.25 & 2.84 \\
    \hline
    MPCL & 8(20\%) & 0 & \textbf{91.98} & \textbf{85.02} & \textbf{4.77} & \textbf{1.58} \\
    \hline
  \end{tabular}}
  \label{fushion}
\end{table}

We also attempt to explore the effects of different prototype fusion steps, which include N-N (no fusion of mixed prototypes), M-L (no fusion of mixed and labeled prototypes), M-UL (no fusion of mixed and unlabeled prototypes), L-UL (no fusion of labeled and unlabeled prototypes), and MPCL (a method combining all prototypes effectively, shown as the best performing setup) within the MPCL framework. The table \ref{fushion} presents the experimental results on LA dataset. As is shown in table \ref{fushion}, N-N performs the worst, indicating the importance of mixed prototypes in enhancing segmentation performance. M-UL and M-L show intermediate performance, suggesting that while the combination of all fusion processes contributes to model performance, the impact of individual fusion is not as critical as the complete fusion process in MPCL. L-UL appears to have an apparent performance degradation, indicating that the global prototypes have better capabilities to optimize hidden embeddings from teacher and student networks in consistency learning, resulting in better segmentation performance.

\begin{table}[htbp]\small
\centering
\caption{Ablation studies of coefficients for prototype fusion on the LA dataset. $\gamma_1 = \frac{\lambda_1}{\lambda_2}$, $\gamma_2 = \frac{\lambda_3}{\lambda_4}$. $\gamma_1$ denotes the fusion ratio of labeled and mixed prototypes. $\gamma_2$ denotes the fusion percentage of unlabeled and mixed prototypes.}
  \setlength{\tabcolsep}{0.8mm}{\begin{tabular}{c|cc|cccc}
    \hline
    \multirow{2}*{$\gamma_1$} &\multicolumn{2}{c|}{Scans Used} &\multicolumn{4}{c}{Metrics}\\
    \cline{2-7}
    {}&{Labeled} & {Unlabled} & Dice$\uparrow$ & Jaccard$\uparrow$ & 95HD$\downarrow$ & ASD$\downarrow$ \\
    \hline
    0.5 & 8(20\%) & 0 & 91.51 & 84.39 & 5.05 & 1.69 \\
    1 & 8(20\%) & 0 & \textbf{91.98} & \textbf{85.02} & \textbf{4.77} & 1.58 \\
    2 & 8(20\%) & 0 & 91.49 & 84.37 & 5.16 & 1.70 \\
    3 & 8(20\%) & 0 & 91.53 & 84.43 & 4.94 & \textbf{1.57} \\
    \hline
    \multirow{2}*{$\gamma_2$} &\multicolumn{2}{c|}{Scans Used} &\multicolumn{4}{c}{Metrics}\\
    \cline{2-7}
    {}&{Labeled} & {Unlabled} & Dice$\uparrow$ & Jaccard$\uparrow$ & 95HD$\downarrow$ & ASD$\downarrow$ \\
    \hline
    0.5 & 8(20\%) & 0 & 91.43 & 84.21 & 5.15 & 1.74 \\
    1 & 8(20\%) & 0 & \textbf{91.98} & \textbf{85.02} & \textbf{4.77} & \textbf{1.58} \\
    2 & 8(20\%) & 0 & 91.65 & 84.57 & 5.02 & 1.66 \\
    3 & 8(20\%) & 0 & 91.54 & 84.32 & 4.98 & 1.62 \\
    \hline
  \end{tabular}}
  \label{coe}
\end{table}

When performing prototype fusion, different fusion ratios may have a significant impact on model performance. Therefore, we also conduct ablation studies to explore the impact of varying coefficients on prototype fusion. The table \ref{coe} displays the experimental results. The experiments involve varying $\gamma_1$ and $\gamma_2$ across four different values (0.5, 1, 2, 3) to observe their impact. Obviously, the optimal value for both $\gamma_1$ and $\gamma_2$ appears to be 1, suggesting that an equal balance among labeled, unlabeled and mixed prototypes is most beneficial for model performance. Increasing $\gamma_2$ from 1 leads to a decrease in performance across all metrics, although the declines are less pronounced compared to changes in $\gamma_1$, indicating that the model might be slightly more sensitive to changes in the balance between labeled and mixed prototypes than between unlabeled and mixed prototypes.

\begin{table}[htbp]\small
\centering
\caption{Ablation studies of the parameter $k$ on the LA dataset. The features from the $k$-rd decoder layer are utilized to generate prototypes. Time represents how many seconds it takes to train a model for one epoch}
  \setlength{\tabcolsep}{0.8mm}{\begin{tabular}{c|cc|cccc|c}
    \hline
    \multirow{2}*{k} &\multicolumn{2}{c|}{Scans Used} &\multicolumn{4}{c|}{Metrics} & \multirow{2}*{Time(s/epoch)}\\
    \cline{2-7}
    {}&{Labeled} & {Unlabeled} & Dice$\uparrow$ & Jaccard$\uparrow$ & 95HD$\downarrow$ & ASD$\downarrow$ & \\
    \hline
    1 & 8(20\%) & 0 & \textbf{91.98} & \textbf{85.02} & \textbf{4.77} & \textbf{1.58} & 38.9 \\
    2 & 8(20\%) & 0 & 91.40 & 84.23 & 5.11 & 1.87 & 35.1\\
    3 & 8(20\%) & 0 & 91.38 & 84.17 & 5.77 & 1.85 & 32.4\\
    4 & 8(20\%) & 0 & 90.74 & 83.15 & 6.18 & 2.00 & 30.5\\
    % 5 & 8(20\%) & 0 & 82.74 & 71.72 & 13.35 & 3.26 \\
    \hline
  \end{tabular}}
  \label{k}
\end{table}

\begin{table}[htbp]\small
\centering
\caption{Ablation studies of consistency loss on the LA dataset. KL-Div denotes the KL divergence loss. MSE denotes the mean square error loss. MAE denotes the mean average error loss. CE represents the cross-entropy loss.}
  \setlength{\tabcolsep}{0.8mm}{\begin{tabular}{c|cc|cccc}
    \hline
    \multirow{2}*{Method} &\multicolumn{2}{c|}{Scans Used} &\multicolumn{4}{c}{Metrics}\\
    \cline{2-7}
    {}&{Labeled} & {Unlabled} & Dice$\uparrow$ & Jaccard$\uparrow$ & 95HD$\downarrow$ & ASD$\downarrow$ \\
    \hline
    KL-Div & 8(20\%) & 0 & 91.50 & 84.45 & 5.17 & 1.73\\
    MSE & 8(20\%) & 0 & 91.31 & 84.02 & 7.32 & 1.85 \\
    MAE & 8(20\%) & 0 & 91.10 & 83.69 & 7.83 & 2.24 \\
     CE & 8(20\%) & 0 & \textbf{91.98} & \textbf{85.02} & \textbf{4.77} &\textbf{1.58} \\
    \hline
  \end{tabular}}
  \label{loss}
\end{table}

Due to the use of trilinear interpolation to upsample prototypes to the same size as segmentation labels, prototypes generated by feature embeddings from different decoder layers may affect the training time and performance of the model. The table \ref{k} shows the results of an ablation study examining the impact of using features from different decoder layers (denoted by the parameter k) for prototype generation. In light of these experimental results, as the value of k increases, there is a consistent decline in model performance across all metrics. This suggests that features from deeper layers, while potentially more abstract or higher-level, may not capture the necessary details for optimal prototype generation. The results also reflect a critical trade-off between segmentation accuracy and training efficiency when selecting which decoder layer's features to use for prototype generation. While utilizing features from the first decoder layer offers the best segmentation performance, it also requires the longest training time at 38.9 seconds per epoch. Conversely, features from deeper layers reduce training time, but at the cost of decreased accuracy and precision. Using features from the fourth layer (k = 4) to generate prototypes is the most time-efficient (30.5 seconds per epoch).

Finally, the table \ref{loss} presents the outcomes of an ablation study examining the effect of different consistency loss functions. According to the results, utilizing cross-entropy loss as consistency loss outperforms those using other loss functions across all metrics, achieving the highest Dice and Jaccard scores (91.98\% and 85.02\%, respectively) and the lowest 95HD and ASD scores (4.77 and 1.58, respectively). However, both MSE and MAE losses result in lower Dice and Jaccard scores compared to KL-Div and CE losses. This could indicate that these loss functions are less suited for capturing the nuances of medical image segmentation tasks.

\section{Conclusions}
The article introduces a novel framework named Mixed Prototype Consistency Learning (MPCL) for semi-supervised medical image segmentation. MPCL integrates a Mean Teacher structure and an auxiliary network to address the limitations of previous prototype-based methods caused by the small quantity and low quality of prototypes. This framework introduces mixed prototypes with additional semantic information to enhance labeled and unlabeled prototypes through prototype fusion, further optimizing the quality and expressiveness of global prototypes, which significantly improve segmentation performance. Extensive experiments on the left atrium and type B aortic dissection datasets demonstrate MPCL's superiority over state-of-the-art approaches, showcasing its effectiveness in medical image segmentation tasks.
\bibliographystyle{splncs04}
\bibliography{main}
\end{document}